%% file: main.tex
\pdfoutput=1
\documentclass[letterpaper, 10 pt, conference]{ieeeconf}  

\IEEEoverridecommandlockouts                              

\usepackage{graphics} 
\usepackage{epsfig} 
\usepackage{amsmath} 
\usepackage{amssymb}  
\usepackage{multirow} 
\usepackage{xcolor}

\usepackage{bm}
\usepackage{booktabs}
\usepackage{multirow}
\usepackage{makecell}
\usepackage{nicefrac}
\usepackage{graphicx}
\usepackage{booktabs}
\usepackage{colortbl}
\usepackage{algorithm}
\usepackage{algorithmic}

\usepackage{graphicx}
\usepackage{subcaption}
\usepackage{comment}
\usepackage{thmtools,thm-restate}
\usepackage{wrapfig}

\makeatletter
\let\NAT@parse\undefined
\makeatother
\usepackage[breaklinks,colorlinks,citecolor=green]{hyperref}
\usepackage{cite}

\title{\LARGE \bf
Controllable Traffic Simulation through LLM-Guided Hierarchical Reasoning and Refinement
}

\author{Zhiyuan Liu$^{1}$, Leheng Li$^{2}$, Yuning Wang$^{1}$, Haotian Lin$^{3}$,\\ Hao Cheng$^{1}$, Zhizhe Liu$^{4}$, Lei He$^{1}$, Jianqiang Wang$^{1, \dagger}$%
\thanks{This work was supported by National key R\&D Program of China: 2023YFB2504402.}
\thanks{$^{1}$ School of Vehicle and Mobility, Tsinghua University.}%
\thanks{$^{2}$ Artificial Intelligence Thrust, The Hong Kong University of Science and Technology (Guangzhou).}%
\thanks{$^{3}$ School of Computer Science, Carnegie Mellon University.}%
\thanks{$^{4}$ Department of Computer Sciences, University of Wisconsin-Madison.}%
\thanks{$^\dagger$ Corresponding author: \tt\footnotesize wjqlws@tsinghua.edu.cn} 
}

\begin{document}

\maketitle
\thispagestyle{empty}
\pagestyle{empty}

\input{parts/01_abstract}
\input{parts/02_intro}
\input{parts/03_related}
\input{parts/04_method}

\input{parts/05_experiments}

\input{parts/06_conclusion}

\bibliographystyle{IEEEtran}
\bibliography{main}

\end{document}

%% file: parts/01_abstract.tex
\begin{abstract}


Evaluating autonomous driving systems in complex and diverse traffic scenarios through controllable simulation is essential to ensure their safety and reliability. However, existing traffic simulation methods face challenges in their controllability. To address this, we propose a novel diffusion-based and LLM-enhanced traffic simulation framework. Our approach incorporates a high-level understanding module and a low-level refinement module, which systematically examines the hierarchical structure of traffic elements, guides LLMs to thoroughly analyze traffic scenario descriptions step by step, and refines the generation by self-reflection, enhancing their understanding of complex situations. Furthermore, we propose a Frenet-frame-based cost function framework that provides LLMs with geometrically meaningful quantities, improving their grasp of spatial relationships in a scenario and enabling more accurate cost function generation. Experiments on the Waymo Open Motion Dataset (WOMD) demonstrate that our method can handle more intricate descriptions and generate a broader range of scenarios in a controllable manner.

\end{abstract}

%% file: parts/02_intro.tex
    \vspace{-2.0mm}

\section{Introduction}\label{sec:intro}

Autonomous driving has developed rapidly in recent years. To ensure the reliability of these systems, testing on complex, diverse, and long-tailed traffic scenarios through traffic simulation is essential and critical.  Early simulation methods, such as rule-based approaches~\cite{8569938, Treiber_2000, rong2020lgsvl}, offered good interpretability but lacked diversity and complexity. In contrast, deep learning techniques \cite{suo2021trafficsim, 10161167,9561666, kamenev2022predictionnet} introduced more complexity by learning from large datasets, but they often lack controllability. Without precise control over generated scenarios, it becomes challenging to systematically evaluate how autonomous driving systems respond to specific, critical situations.


To address this problem, recent work has focused on utilizing generative models to achieve controllable traffic simulation. One approach \cite{pronovost2023scenario, rowe2024ctrlsimreactivecontrollabledriving, wang2024dragtrafficnonexpertinteractivepointbased} involves using condition tokens to guide the model with basic conditions, such as goal points or velocity. However, this method requires additional training with the condition tokens and struggles to handle more complex conditions. Another approach leverages conditional sampling in diffusion models, which does not require extra training and offers more flexibility by optimizing cost functions during the sampling process. The cost functions can be either human-designed \cite{zhong2023guided, jiang2023motiondiffuser, chang2023controllable, xu2023diffscene} or generated by large language models (LLMs) \cite{zhong2023language, jiang2024scenediffuser}. Human-designed cost functions typically include predefined rules, such as speed limits or avoiding collisions, but are limited in their ability to address conditions outside of predefined functions. On the other hand, methods based on LLM-generated cost functions mark a significant progression to deal with a wide range of conditions. They take user descriptions of a scenario as input and create corresponding cost functions by LLMs, which is more adaptable than predefined functions. However, current LLMs still face challenges in either processing intricate descriptions effectively or generating outputs that are easily guidable, resulting in a low success rate for producing correct traffic simulation.

Our goal is to overcome the limitations of simulation methods based on diffusion models and LLM-generated cost functions, further utilizing the powerful generative capabilities of LLMs and leveraging the flexibility of diffusion conditional sampling. To achieve this, we propose a novel LLM-based framework consisting of two key modules: a high-level understanding module, and a low-level refinement module.

The high-level module is designed to guide LLMs in processing complex conditions and generating appropriate cost functions. Here, we systematically design a novel chain-of-thought (CoT) mechanism~\cite{wei2023chainofthoughtpromptingelicitsreasoning} for traffic scene analysis, which enables LLMs to systematically break down the traffic events into hierarchical structures and generate corresponding cost functions step-by-step. By reasoning the interaction graphs in the traffic scene and breaking them down into manageable components, we demonstrate the significant advantages in understanding and processing complex traffic scenes. However, the generated cost functions may occasionally be suboptimal or challenging to control via diffusion guidance. To address this, we introduce a low-level refinement module, which takes the low-level trajectory coordinates as inputs and enables the LLM to perform self-reflection and iterative improvement. This module further refines the generated cost functions, enhancing the model's ability to comprehend complex scenes and improving the performance of guided simulation.

Moreover, the formulation of appropriate cost functions is crucial. Existing approaches often rely on target points \cite{jiang2024scenediffuser} or project trajectories onto surrounding lanes and utilize the projection points \cite{zhong2023language} to construct cost functions. However, these methods present significant limitations. The use of target points overly restricts the feasible trajectory space, as a single description may correspond to multiple valid target points rather than a single one. Additionally, the geometric interpretation of projection points is inherently difficult for LLMs to comprehend, and the process of translating these projections into cost functions further complicates the task. To address these issues, we propose a novel Frenet-frame-based cost function framework. This approach leverages lateral and longitudinal information to construct cost functions, providing clear geometric interpretations and more reasonable constraints. We demonstrate that this Frenet-based method not only enhances the LLM’s understanding of the scenario but also significantly improves the success rate of diffusion guidance.


Regarding the model architecture, we integrate the state-of-the-art trajectory prediction frameworks \cite{zhou2023query, lan2024sept, lin2024joint} with a relative coordinate system to ensure translational and rotational invariance for each agent in the scene. Furthermore, we formulate the agent's known trajectories as a cost function that serves as guidance during diffusion test-time sampling. This design provides a flexible framework that can accommodate various applications, including unconditional generation, and conditional simulation, without relying on extra training and complex models.

We evaluate our method on the Waymo Open Motion Dataset (WOMD) \cite{Ettinger_2021_ICCV}. Experimental results demonstrate that the proposed LLM framework and Frenet-based cost functions can effectively handle a wider range of driving scenarios compared to existing methods. This includes not only typical driving behaviors, such as cut-ins, but also abnormal behaviors, such as going off-road. Additionally, our method achieves zero-shot ability on accident events despite being trained solely on a nearly collision-free dataset. Extensive experiments further validate that our approach delivers competitive performance across multiple tasks.

Our contributions are summarized as follows:

\begin{itemize}
    \item We propose a novel LLM-based framework designed to interpret complex traffic scenarios and generate precise cost functions, enabling more accurate and reliable traffic simulation.
    
    \item We innovatively address the effectiveness of Frenet coordinate system in traffic simulation domain, significantly enhancing the generative capabilities of LLMs and improving the success rate of diffusion-guided scenario generation.

    \item Through extensive experiments, we demonstrate that our method can handle more intricate scenario descriptions and generate a wider variety of traffic scenarios in a controllable manner, including complex interactions and accident simulations.

\end{itemize}



%% file: parts/03_related.tex
\section{Related Work}\label{sec:related}

\subsection{Traffic simulation}

\textbf{Rule-based methods}. Early methods utilize heuristic controllers or replayed logs to generate scenarios and use human-designed rules to control the generation \cite{8569938, Treiber_2000, rong2020lgsvl}. These methods have good controllability and interpretability. However, they lacked the realism and complexity to capture the full range of real-world situations.

\textbf{Learning-based methods}. Learning-based simulation methods leverage large-scale real-world datasets to simulate driving behaviors \cite{Ettinger_2021_ICCV, chang2019argoverse, wilson2023argoverse, nuscenes2019}. Early models in this domain \cite{suo2021trafficsim, 10161167,9561666, kamenev2022predictionnet} were designed similarly to trajectory prediction models \cite{varadarajan2022multipath++, gao2020vectornet} but operated in a closed-loop manner. While these methods provide improved realism, diversity, and complexity, they often lack sufficient controllability. To address this, some approaches incorporate conditional models \cite{pronovost2023scenario, rowe2024ctrlsimreactivecontrollabledriving, wang2024dragtrafficnonexpertinteractivepointbased, zhong2023guided, jiang2023motiondiffuser, chang2023controllable, xu2023diffscene, zhong2023language} to balance both realism and controllability.

\textbf{Diffusion-based methods}. More recent work leverages diffusion models for controllable simulation, benefiting from strong generative capabilities and test-time adaptability. Most of these models \cite{zhong2023guided, jiang2023motiondiffuser, chang2023controllable, xu2023diffscene, huang2024versatile} rely on pre-defined functions, such as goal point specification and collision avoidance, to guide the sampling process. While effective, their ability to control conditions beyond these predefined functions is constrained.  CTG++ \cite{zhong2023language} and SceneDiffuser \cite{jiang2024scenediffuser} marks a significant progression by utilizing large language models (LLMs) to translate user-specified descriptions into cost functions, which expands the range of controllable conditions. However, it still faces challenges in understanding and processing complex descriptions.

\subsection{LLMs for robotics}

The application of LLMs in robotics has garnered significant attention due to their ability to bridge the gap between natural language instructions and robotic control. Recent approaches leverage LLMs to generate both actions \cite{10160396, 10182264} and reward functions \cite{lin-etal-2022-inferring, yu2023language, kim2024llmbased}, achieving notable progress in this area. Building on these advancements, CTG++ \cite{zhong2023language} and SceneDiffuser \cite{jiang2024scenediffuser} explored the use of LLMs in traffic simulation, employing them to generate cost functions based on natural language descriptions of traffic scenarios. In this work, we aim to extend the generative capabilities of LLMs to construct more comprehensive and precise cost functions, enhancing the controllability of traffic simulations.

%% file: parts/04_method.tex
\section{Preliminary}\label{sec:preliminary}

\begin{figure*}
    \centering
    \includegraphics[width = 0.92\textwidth]{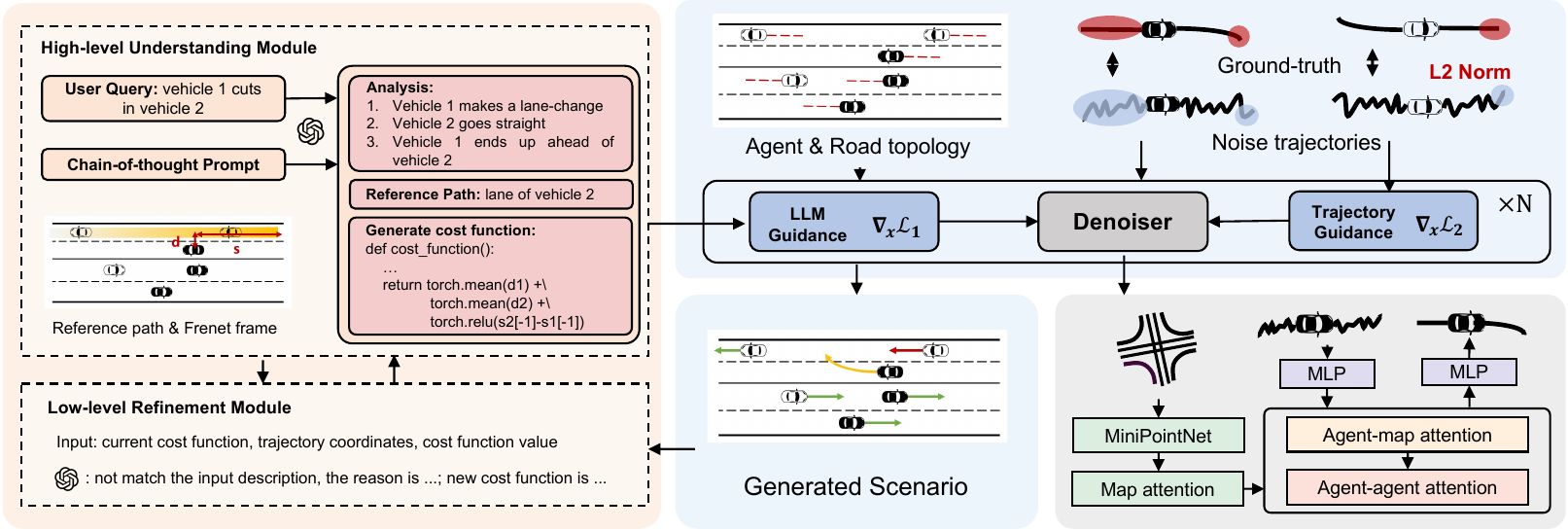}
    
    \caption{
        \textbf{Method Overview}. We employ LLM to generate guidance and utilize a diffusion model to produce trajectories. The high-level module generates a cost function, which guides the transformer-based diffusion model to generate trajectories. If the trajectory generation fails, the coordinates are fed into the low-level module to further refine the cost function.
    }
    \label{fig:pipeline}
    \vspace{-6.0mm}
\end{figure*}

\subsection{Problem formulation}

A traffic scenario can be defined as a combination of road topology $m$ and agent states $a$. We use a set of $N_{m}$ polylines for road topology to describe lanes~\cite{gao2020vectornet}, characterized by their coordinates and attributes (such as lane type). In addition, we denote the connectivity of the polylines (predecessor, successor, left neighbor, and right neighbor) as graph edges $E=\{e_{ij}\}$, to represent the road topology. For agent states, we focus primarily on the position and heading of each agent over a sequence of timesteps. An agent’s state history and future are denoted as $a_i = \left\{\boldsymbol{s}_{-T_{\text{hist}}}, ..., \boldsymbol{s}_0, ..., \boldsymbol{s}_{T_{\text{future}}}\right\}$, where $\boldsymbol{s}_t = (x_t, y_t, \theta_t)$. Here, $T_{\text{hist}}$ and $T_{\text{future}}$ represent the lengths of the history and future trajectories, respectively.

Our goal is to simulate $N_a$ agent future trajectories given (1) the road topology $m$, (2) the current or historical states of $N_a$ agents, and (3) the language description specified by the user. These generated trajectories must be consistent with the scenario's language description.

\subsection{Diffusion Models}
\label{diffusion}

Diffusion models~\cite{ho2020denoising} have emerged as a powerful framework to represent complex data distribution. In general, diffusion models parameterize data distribution as $p_{\theta}$ and learn to model it through score-matching, transforming a simple distribution (e.g., Gaussian noise) into a complex data distribution via multiple small, reversible steps. This process is modeled by stochastic differential equations (SDEs)~\cite{song2020score}:

\begin{equation}
\label{eqn: forward sde}
    d\boldsymbol{x} = f(\boldsymbol{x},t)dt+g(t)d\omega
\end{equation}
\begin{equation}
\label{eqn: backward sde}
    d\boldsymbol{x} = [f(\boldsymbol{x},t)-g^2(t)\nabla_{\boldsymbol{x}}{\log{p_t}(\boldsymbol{x})}]dt+g(t)d\omega
\end{equation}
where $\omega$ is the standard Wiener process. $\nabla_{x}{\log{p_t}(x)}$ is the time-dependent score, which can be approximated by a neural network. In our case, diffusion models are used to represent the distribution of trajectory features $x$ conditioned on map topology $p_{\theta}(x; m)$.

Following EDM~\cite{karras2022EDM}, the sampling process in this work is described by the ODE:
\begin{equation}
    \label{eqn: edm_sim_ode}
    d\boldsymbol{x} = - \dot{\sigma}(t)\sigma(t) \nabla_x \log p(\boldsymbol{x};m,\sigma(t)) dt
\end{equation}

The denoiser model $D_{\theta}(\boldsymbol{x}; m, \sigma)$ is trained to recover the noised sample $\boldsymbol{x}$ from its noisy version. The relationship between the score function and the denoiser model is expressed as follows:
\begin{equation}
\label{eqn: score_and_denoiser}
    \nabla_{\boldsymbol{x}} \log p_{\theta}(\boldsymbol{x}; m, \sigma)=\frac{D_{\theta}(\boldsymbol{x}; m,\sigma) - \boldsymbol{x}}{\sigma^2}
\end{equation}
During training, the model is optimized by minimizing the following objective:
\begin{equation}
    \mathbb{E}_{y \sim p_0, \boldsymbol{n}\sim\mathcal{N}(\boldsymbol{0}, \sigma ^2 \boldsymbol{I})} \left\Vert D_{\theta} (\boldsymbol{y} + \boldsymbol{n} ; m, \sigma) - \boldsymbol{y} \right\Vert^2,
\end{equation}
where $p_0$ is the data distribution and $\boldsymbol{n}$ represents the Gaussian noise. After training, data can be sampled from noise using the relationship in equation (\ref{eqn: score_and_denoiser}) and the ODE described in equation (\ref{eqn: edm_sim_ode}).


\subsection{Conditional sampling}
\label{conditional}
For various problems, we have extra condition $c$ which can be described by a cost function $\mathcal{L}$. In such cases, we aim to sample data from the conditional distribution $p(\boldsymbol{x};c)$. Under our specific problem settings, this translates to sampling from $p(\boldsymbol{x};m,\sigma,c)$. Consequently, the score function should be modified from $\nabla_{\boldsymbol{x}}{\log{p}(\boldsymbol{x};m,\sigma)}$ to $\nabla_{\boldsymbol{x}}{\log{p_t}(\boldsymbol{x};m,\sigma,c)}$. According to  Bayes’ rule, the modified score function $\nabla_{\boldsymbol{x}}{\log{p}(\boldsymbol{x};c,m,\sigma)}$ can be expressed as:
\begin{equation}
     \nabla_{\boldsymbol{x}}{\log{p}(\boldsymbol{x};m,\sigma)} + \nabla_{\boldsymbol{x}}{\log{p}(c;\boldsymbol{x},m,\sigma)}
\end{equation}
The first term can be calculated by equation (\ref{eqn: score_and_denoiser}), and the second term can be approximated as:
\begin{equation}
    \label{eqn: guidance score}
    \nabla_{\boldsymbol{x}}{\log{p}(c;\boldsymbol{x},m,\sigma)} \approx \lambda \frac{\partial }{\partial \boldsymbol{x}} \mathcal{L} (D_{\theta} (\boldsymbol{x}; m, \sigma)),
\end{equation}
where $\lambda$ is a scaling factor. By utilizing equations (\ref{eqn: edm_sim_ode}), (\ref{eqn: score_and_denoiser}) and (\ref{eqn: guidance score}), we can sample data that meets the condition $c$. In practice, to ensure the stability of the sampling process, a score threshold of $(-1, 1)$ is applied to equation (\ref{eqn: guidance score}).

\section{Method}
\label{sec:method}

The overall framework of our proposed method is illustrated in Figure~\ref{fig:pipeline}. In this section, we first introduce the \textbf{Transformer-based Denoiser}, which models the complex interactions within traffic scenarios (see Section~\ref{sec:denoiser}). Following this, we detail the LLM-based \textbf{Guided Generation} module in Section~\ref{sec:generation}.

\subsection{Transformer-based denoiser}
\label{sec:denoiser}
\textbf{Relative coordinate system}.  In alignment with recent trajectory prediction methods~\cite{zhou2023query, shi2023mtr}, we adopt a relative coordinate system. Initially, road polylines and trajectories are rotated and normalized to the central or current position. To further capture spatial relationships among traffic elements, the feature vector $\boldsymbol{d}_{ij} = (\Delta x_{ij}, \Delta y_{ij}, \cos \Delta \theta_{ij}, \sin \Delta \theta_{ij})$ is extracted, where $(\Delta x_{ij}, \Delta y_{ij})$ represents the relative positions within the reference frame of agent $i$, and $\Delta \theta_{ij}$ denotes the relative heading angle. Note that the feature vectors between static elements, such as road polylines, remain consistent regardless of the agents' movements. Consequently, this approach eliminates the need to re-encode the map topology during simulation rollout, thereby enhancing inference speed.



\textbf{Trajectory conditional guidance}. Our approach treats the known trajectory states $A_{\text{known}}$ as an additional condition rather than embedding it directly into the model. Consequently, our model learns the distribution $p(x; m)$ and employs the sampling method introduced in Section~\ref{conditional} to generate data conditioned on historical states. Let $\mathcal{T} = \{t_1, t_2, \dots, t_k\}$ denote the set of timesteps where condition states are provided. The predicted trajectory states at these timesteps are represented as $\boldsymbol{y}_{\text{known}} = \{\boldsymbol{y}_{t_i} \mid t_i \in \mathcal{T}\}$ , and the corresponding ground-truth states are $\hat{\boldsymbol{y}}_{\text{known}} = \{\hat{\boldsymbol{y}}_{t_i} \mid t_i \in \mathcal{T}\}$. The condition is expressed using the L2 norm:
\begin{equation}
\mathcal{L}_{\text{known}}=\left\| \boldsymbol{y}_{\text{known}} - \hat{\boldsymbol{y}}_{\text{known}} \right\| ^2
\end{equation}
 By incorporating this loss term, we effectively integrate trajectory conditions without requiring separate encoders and additional training. This design enables our method to flexibly handle various tasks, such as unconditional prediction and conditional simulation.

\textbf{Model architecture}. The model architecture is shown in the bottom right of Figure~\ref{fig:pipeline}. We employ a MiniPointNet~\cite{qi2017pointnet} which is primarily composed of max pooling and MLPs to obtain polyline-level features from the road topology feature $M$. The encoding process can be expressed as: 
\begin{equation}
    X_m=\mathrm{MiniPointNet}(M), X_m \in \mathbb{R}^{N_m \times d}\label{4-1}
\end{equation}
where $X_m$ is the polyline-level feature and $d$ denotes the hidden size of the model. For agent features $X_a \in \mathbb{R}^{N_a \times d}$, we embed the trajectories into PCA space following \cite{jiang2023motiondiffuser} and then utilize an MLP to encode the noise trajectories.

After feature encoding, we employ three distinct attention modules to extract the lane-lane, agent-lane, and agent-agent interactions. Let $\boldsymbol{x}_i$ denote the feature of an object, the attention mechanism can be expressed as: 
\begin{align}
    \tilde{\boldsymbol{x}_i}=&\sum_{j} \mathrm{Softmax}(s_{ij})(\boldsymbol{x}_j \boldsymbol{W}_v + \mathrm{MLP}(\boldsymbol{r}_{ij})), \\
    s_{ij}=&\frac{1}{\sqrt{d}}(\boldsymbol{x}_i \boldsymbol{W}_q)(\boldsymbol{x}_j \boldsymbol{W}_k + \mathrm{MLP}(\boldsymbol{r_{ij}})),\label{4-2}
\end{align}
where $\boldsymbol{W}$ represents a Linear layer. For agent-agent and agent-lane attentions, $\boldsymbol{r}_{ij}$ is the spatial relativity feature $\boldsymbol{d}_{ij}$. For lane-lane attentions, $\boldsymbol{r}_{ij}=\boldsymbol{d}_{ij}+\boldsymbol{e}_{ij}$ is the sum of the spatial relativity feature and the connectivity feature.

\subsection{Guided generation}
\label{sec:generation}

Our goal is to convert user descriptions into cost functions using LLMs and guide the generation through the method outlined in~\ref{conditional}. Existing approaches often face challenges in processing and comprehending complex descriptions. To address this issue, we propose an LLM-based framework comprising a high-level understanding module and a low-level refinement module.
 
\begin{figure}
    \centering
    \includegraphics[width = 0.45\textwidth]{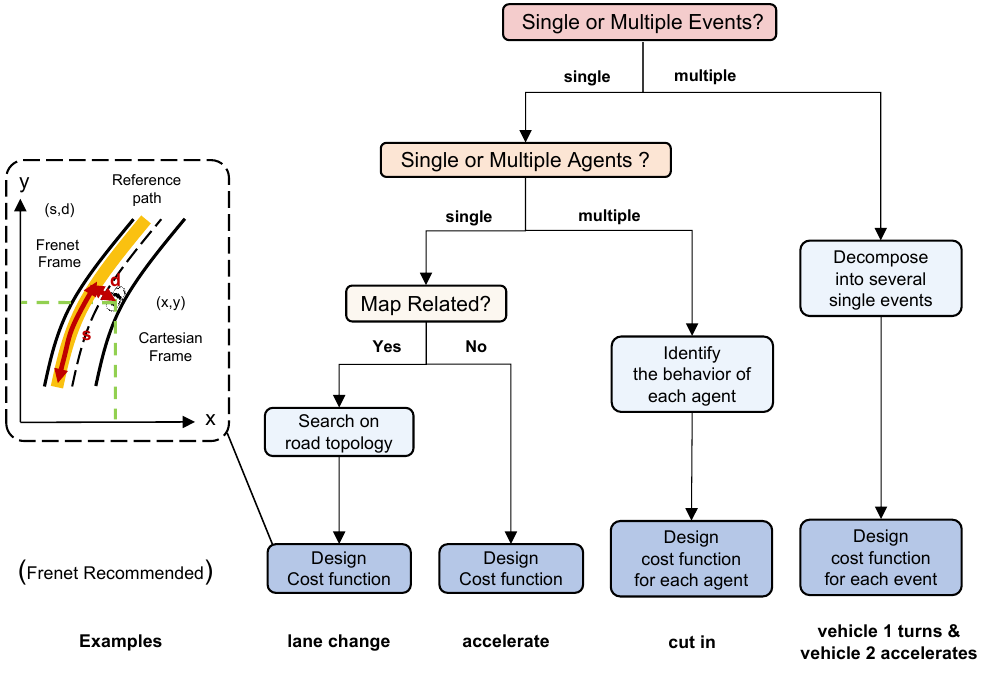}
    \caption{
    We systematically design the chain-of-thought mechanism to thoroughly analyze and interpret the hierarchical behaviors in traffic scenes for controllable generation.
    }
    \label{fig:CoT}
    \vspace{-6.0mm}
\end{figure}

\textbf{High-Level Understanding Module}. The high-level module employs a systematically designed Chain-of-Thought (CoT) mechanism to reason about the interaction graph and decompose complex events into manageable components. Figure~\ref{fig:CoT} illustrates the detailed CoT process. First, the module identifies the events. For multiple events, it breaks them down into individual single events. Next, it determines the number of agents involved. If multiple agents are present, it identifies each agent's behavior. Subsequently, it evaluates whether each behavior is map-related. If so, it designs an algorithm to explore the road topology $E=\{e_{ij}\}$. Finally, it generates cost functions. For map-related descriptions, Frenet-based cost functions are recommended. We demonstrate that this structured reasoning approach enables the LLM to effectively understand and process a broader range of complex descriptions.

\textbf{Low-Level Refinement Module}. While the high-level module excels at generating accurate cost functions for complex descriptions, the resulting cost functions may occasionally be suboptimal or challenging to control through diffusion guidance. To address this, we introduce the low-level refinement module to further refine the cost functions. This module takes the low-level generated trajectory coordinates as input, enabling the LLM to directly analyze the generated trajectory data. Additionally, the corresponding cost function values from the trajectory optimization process are provided, offering further insights for the LLM's analysis. Using these inputs, the LLM evaluates whether the trajectory aligns with the given description, whether the cost function accurately reflects the desired behavior, and whether each loss term has appropriate hyperparameters. For example, the LLM can directly assess the correctness of the behavior from the trajectory coordinates and identify potential issues from unusually high cost values, enabling precise error analysis and correction. Through this design, we enhance the framework’s ability to understand complex traffic scenarios and dynamically adapt to varying traffic contexts.

\textbf{Frenet-based cost functions}. Inspired by the success of the Frenet frame in motion planning, we utilize it to construct cost functions for descriptions related to map information. The Frenet frame is shown on the left side of Figure~\ref{fig:CoT}. It describes the relative position of trajectories concerning the reference path by the longitudinal distance $s$ and the lateral offset $d$. These parameters have clear geometric interpretations, making them intuitive for LLMs to understand and utilize. Furthermore, unlike methods that rely on target points, which can overly constrain the feasible trajectory space, our Frenet-based cost functions provide more reasonable constraints by naturally decomposing trajectories into longitudinal and lateral components. LLMs will generate cost functions based on $s$ and $d$, enabling the control of both longitudinal and lateral directions. We will demonstrate that these Frenet-based cost functions significantly improve the LLMs' understanding of spatial relationships in traffic scenarios, leading to more accurate cost function generation and higher guidance success rates.

%% file: parts/05_experiments.tex
\begin{table}[tb]
\centering
\resizebox{0.50\textwidth}{!}{
\begin{tabular}{@{}lccccc@{}}
\toprule
\multirow{2}{*}{\textbf{Method}} & ADE & FDE  & JSD & Collision & Off Road \\
& (m) $\downarrow$ & (m) $\downarrow$  & ($\times 10^{-2}$) $\downarrow$ & (\%) $\downarrow$ & (\%) $\downarrow$ \\
\midrule
 Actions-Only \cite{philion2024trajeglish} & $4.81$ & $11.89$ & $10.4$ & $19.9$ & $27.6$ \\
 DT \cite{NEURIPS2021_7f489f64} & $1.56$ & $3.07$ & $8.4$ & $\underline{5.3}$ & $\mathbf{11.0}$ \\
 CtRL-Sim \cite{rowe2024ctrlsimreactivecontrollabledriving} & $\underline{1.25}$ & $\underline{2.04}$ & $\underline{7.9}$ & $\underline{5.3}$ & $\mathbf{11.0}$ \\
\midrule
CTG++ \cite{zhong2023language} & $1.73$ & $4.02$ & $\mathbf{7.4}$ & $5.9$ & $15.0$ \\
\textbf{Ours} & $\mathbf{1.17}$ & $\mathbf{1.93}$ & $8.2$ & $\mathbf{5.1}$ & $\underline{12.8}$\\
\bottomrule
\end{tabular} 
}
\caption{Results of conditional simulation. We \textbf{bolded} the optimal result and \underline{underlined} the second best result.}
\label{tab:sim_results}
\vspace{-6.0mm}
\end{table}

\section{Experiments}\label{sec:experiments}
    \vspace{-1.0mm}

\subsection{Experiment Setup}

\textbf{Dataset}. Our experiments are conducted on large-scale real-world traffic scenario dataset: the Waymo Open Motion Dataset (WOMD)~\cite{Ettinger_2021_ICCV}. WOMD comprises over 570 hours of 10hz scenario data. We train our model on the entire training split, which contains  69,500 scenarios and evaluate it on a subset of 1,000 scenarios from validation split.

\textbf{Evaluation}. We begin by evaluating our model design through experiments on two tasks. The first task is \textbf{conditional simulation}, where we follow the experimental setup of CtRL-Sim \cite{rowe2024ctrlsimreactivecontrollabledriving}. In this setup, the model is provided with a 1.1-second history and an endpoint, and it simulates future trajectories over an 8-second horizon.  We use the Final Displacement Error (\textbf{FDE}) and Average Displacement Error (\textbf{ADE}) to measure reconstruction error. We employ a distributional metric (\textbf{JSD}) to evaluate realism, defined as the mean of the Jensen-Shannon Distances computed on linear speed, angular speed, acceleration, and distance to the nearest vehicle between real and simulated scenes. Additionally, common-sense metrics like \textbf{Collision} rate and \textbf{Offroad} rate are used.

The second task is \textbf{unconditional generation}, where only the current states are provided. We evaluate the model using scene collision rate (\textbf{SCR}) and the Maximum Mean Discrepancy (\textbf{MMD}) metrics for both distance to the nearest object features ($\mathrm{MMD_o}$) and distance to road edge features ($\mathrm{MMD_r}$).

Finally, we conduct experiments on \textbf{guided simulation}. First, we select a range of scenario descriptions and assess whether the generated cost functions accurately reflect these descriptions. Next, we further choose four types of events, select 20 scenarios for each type, and evaluate the generation success rate. We visualize the results to determine whether the model correctly adheres to the guidance provided by the generated cost functions in each scenario.

\begin{table}[t]
\centering
\resizebox{0.4\textwidth}{!}{
\begin{tabular}{lccc}
    \toprule
    \textbf{Method} & SCR(\%)$\downarrow$ & $\mathrm{MMD_o}\downarrow$ &  $\mathrm{MMD_r}\downarrow$  \\
    \midrule
     TrafficGen \cite{feng2023trafficgen} & 9.7 & 0.31 & 0.23 \\
     LCTGen \cite{tan2023language} & 11.2 & 0.27 & 0.33 \\
     \textbf{Ours} & \textbf{6.3} & \textbf{0.24} & \textbf{0.17} \\
    \bottomrule
\end{tabular}}
\caption{Results of unconditional scenario generation.}
\label{tab:generation_compare}
\vspace{-6.0mm}
\end{table}

\textbf{Baselines}. For conditional simulation, we compare our method against recent auto-regressive methods, including Actions-Only (a modified version of Trajeglish \cite{philion2024trajeglish}), Decision Transformer \cite{NEURIPS2021_7f489f64}, and CtRL-Sim \cite{rowe2024ctrlsimreactivecontrollabledriving}. We also include the popular diffusion-based method CTG++ \cite{zhong2023language}. For unconditional generation, we compare our approach to TrafficGen \cite{feng2023trafficgen} and LCTGen \cite{tan2023language}. In the guided generation experiments, we compare our proposed LLM-based framework in comparison to CTG++ \cite{zhong2023language} and SceneDiffuser \cite{jiang2024scenediffuser}.

\textbf{Implementation Details}. We set the number of layers for all attention modules to 3. The number of agents is set to $N_a = 24$, and the number of lane polylines to $N_m = 200$. We use 32 sampling steps for diffusion sampling, with other hyperparameters following EDM \cite{karras2022EDM}. For closed-loop simulation, we set the re-plan rate to 2 Hz. For guided simulation, we use GPT-4o \cite{achiam2023gpt} for cost function generation.

\begin{table*}[ht]
\centering

\resizebox{0.9\textwidth}{!}{
\begin{tabular}{clcccccc}
    \toprule
    \multirow{2}{*}{\textbf{Category}} & \multicolumn{1}{c}{\multirow{2}{*}{\textbf{\ Description}}} & \multirow{2}{*}{ CTG++} & \multirowcell{2}{ Scene \\ Diffuser} & \multirowcell{2}{ Ours \\ w/o understanding} & \multirowcell{2}{ Ours \\ zero-shot CoT} &
    \multirowcell{2}{ Ours \\ w/o Frenet} &
    \multirow{2}{*}{Ours}\\
    & & &  & & & &\\
    \midrule
     \multirowcell{3}{ Basic \\  rules} &  vehicle drives at a speed limit 10m/s & \checkmark & \checkmark & \checkmark & \checkmark & \checkmark & \checkmark\\
     &  vehicle reaches target point (25, 25) & \checkmark & \checkmark & \checkmark & \checkmark & \checkmark & \checkmark \\
     &  no collisions between all vehicles & \checkmark & \checkmark & \checkmark & \checkmark & \checkmark & \checkmark \\

    \midrule

    \multirowcell{3}{ Behavior and \\  interactions} &  vehicle makes a left-lane-change & \checkmark & & \checkmark & \checkmark & \checkmark & \checkmark\\
     &  vehicle moves to the rightmost lane &  & &  &  & \checkmark & \checkmark \\
     &  vehicle 1 cuts in vehicle 2 &  & \checkmark & & \checkmark & & \checkmark\\

    \midrule

    \multirowcell{3}{ Abnormal \\  behaviors} &  vehicle drives in reverse & & & \checkmark & & & \checkmark \\
     &  vehicle drives out of the road & & & & & & \checkmark \\
     &  vehicle drives left and right within the lane & & \checkmark &  &  &  & \checkmark \\

    \bottomrule
\end{tabular}}
\caption{Results of controllable simulation.}
\label{tab:controllable_compare}
\vspace{-6.0mm}
\end{table*}

\subsection{Traffic simulation results}

The conditional simulation results are presented in Table~\ref{tab:sim_results}. Our model achieves the best reconstruction performance, as measured by FDE and ADE, and ranks second in terms of common-sense metrics such as Collision Rate and Offroad Rate. However, the Jensen-Shannon Distance (JSD) metric is slightly inferior to the current state-of-the-art methods.  A potential reason for this is that our method directly outputs future states, whereas other methods generate actions, which may lead to better dynamic feasibility.

The results of unconditional generation, shown in Table~\ref{tab:generation_compare}, further demonstrate the effectiveness of our approach. Our model outperforms all baselines across metrics such as Scene Collision Rate (SCR) and Maximum Mean Discrepancy (MMD), confirming its ability to generate realistic traffic scenarios. These two experiments show that our flexible model design can handle various tasks and achieve competitive performance.

\subsection{Guided simulation results}

\textbf{Evaluation on generated cost functions}. Firstly, we evaluate our method using a diverse set of scenario descriptions, including basic rules, common behaviors and interactions, and abnormal behaviors, as detailed in Table~\ref{tab:controllable_compare}. We analyze the generated cost functions to verify their alignment with the input descriptions. To validate the contribution of each component in our method, we conduct an ablation study with the following variations: (1) our method without the high-level understanding module, and (2) our method without the Frenet coordinate system. Additionally, to further demonstrate the effectiveness of our design, we replace our CoT with zero-shot CoT \cite{NEURIPS2022_8bb0d291}, where the LLM is instructed to "think step by step" and reason about the traffic event independently, and compare the results.

\begin{table}[t]
\centering

\resizebox{0.45\textwidth}{!}{
\begin{tabular}{cccc}
    \toprule
    \multirow{2}{*}{\textbf{Scenario Type}} & \multirowcell{2}{ Scene \\ Diffuser} &
    \multirowcell{2}{ Ours \\ w/o Refine} &
    \multirowcell{2}{Ours \\ w/ Refine}\\
     & & &\\
    \midrule
    Cut in & 0.45 & 0.65 & 0.85 \\
    Out of road & - & 0.70 & 0.90 \\
    Yield & 0.40 & 0.70 & 0.85 \\
    Rightmost & - & 0.75 & 0.95 \\
    \bottomrule
\end{tabular}
}
\caption{Results of guidance success rate.}
\label{tab:succ_rate}
\vspace{-6.0mm}
\end{table}

In terms of basic rules, all tested methods demonstrated effective performance by employing functions like Norm for target point and ReLU for speed limit. These results suggest that LLMs do not confine the generation of cost functions to recommended forms (e.g., Frenet), but instead allow for diverse, adaptable cost function formulations.

When addressing behaviors such as lane changes, all methods performed well. However, for more complex road topologies such as navigating the rightmost lane, methods with our designed CoT demonstrated superior performance due to their enhanced map exploration capabilities. In agent interactions like cut-ins, our understanding module effectively understands and processes them by recognizing that one vehicle continues straight while another performs a lane change into its path, while CTG++ struggles to understand them. Although correctly understanding these behaviors, methods without Frenet coordinates cannot sufficiently understand the spatial relationships thus generating wrong functions. In contrast, the Frenet-based approach performs accurately, as its clear geometric interpretations further enhance the LLM’s understanding of the scenario.



Regarding abnormal behaviors, our approach demonstrates impressive performance. The LLM generates $sin(d)$ to simulate the vehicle swaying left and right within the lane,  and controls $d$ to be positive, ensuring the vehicle stays on the left side of the road edge. The Frenet-based approach allows the LLM to express these abnormal behaviors with concise function forms, highlighting the expressiveness of the Frenet frame and its ability to enhance the LLM’s understanding of complex traffic scenarios. 

Through this experiment, we demonstrate that our method can effectively handle a broader range of scene descriptions compared to current approaches. Furthermore, when compared to zero-shot CoT, our method exhibits significantly stronger reasoning capabilities, highlighting that understanding complex traffic scenes requires more than just the inherent knowledge of LLMs. Instead, it necessitates a systematically designed framework to interpret and reason about traffic interaction graphs. Additionally, the integration of the Frenet coordinate system proves to be a critical component, enabling more precise spatial reasoning and accurate generation.

\begin{figure}[t]
    \centering
    \includegraphics[width = 0.45\textwidth]{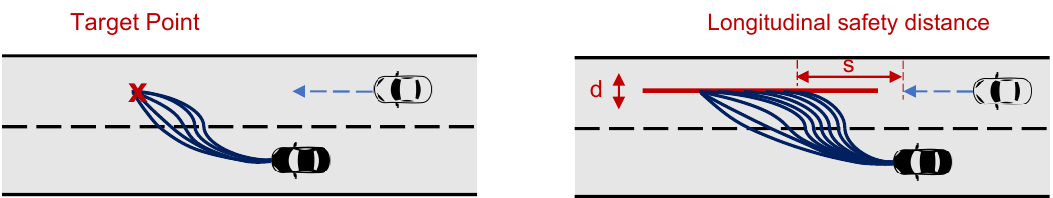}
    
    \caption{
    Visualization of constraints. Compared to reaching the same endpoint, our method provides a more reasonable constraint, making it easier for the diffusion to follow.
    }
    \label{fig:tp_vs_frenet}
    \vspace{-4.0mm}
\end{figure}

\textbf{Evaluation on success rate of guided simulation}. To further validate the correctness of our LLM generation, we conduct a guidance success rate experiment across four types of scenarios, comparing our approach with a baseline method based on target points. The results, as shown in Table~\ref{tab:succ_rate}, demonstrate that our method achieves a significantly higher success rate than the target point-based baseline. This improvement can be attributed to the design of our Frenet-based cost functions, which provide reasonable and flexible constraints, as illustrated in Figure \ref{fig:tp_vs_frenet}.

\begin{figure}
    \centering
    \includegraphics[width = 0.45\textwidth]{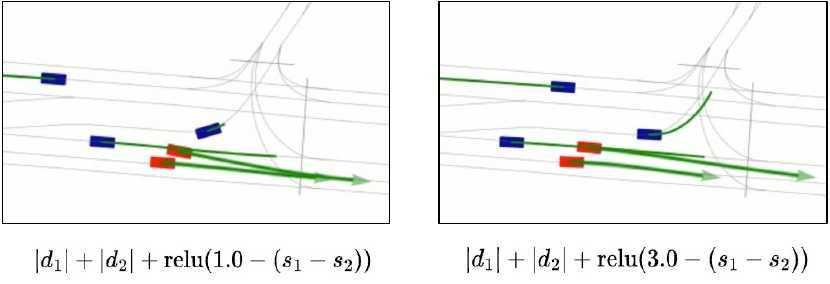}
    
    \caption{
    A cut-in example of refinement: Two vehicles collide with each other and the longitudinal loss term is not effective. We should increase the safety distance in the cost function.
    }
    \label{fig:refinement}
    \vspace{-6.0mm}
\end{figure}

\begin{figure*}
\centering
\subfloat[][\small{User query: vehicle 1 cuts in vehicle 2.}]{\includegraphics[width=0.47\textwidth,trim={0 0 0 0.03cm},clip]{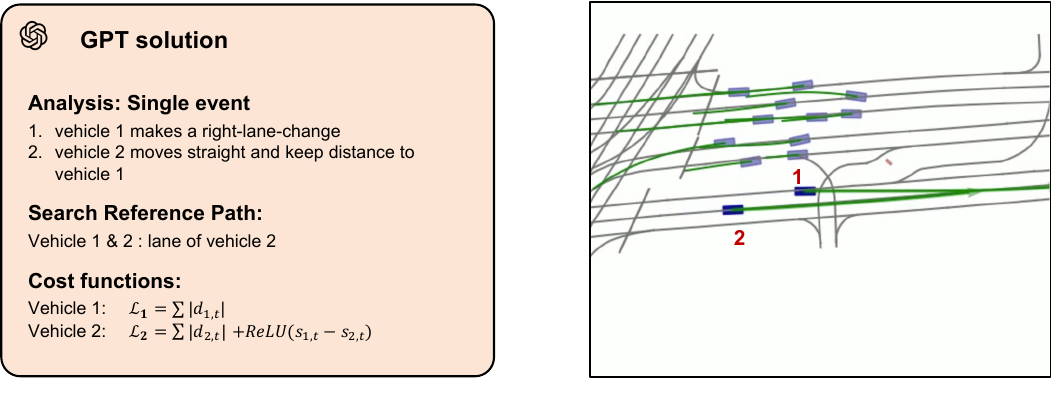}}
\hspace{0.03\linewidth}
\subfloat[][\small{User query: vehicle 1 moves to the rightmost, while vehicle 2 drives straight and vehicle 3 follows vehicle 2.}]{\includegraphics[width=0.47\textwidth, trim={0 15.0 0 0cm}, clip]{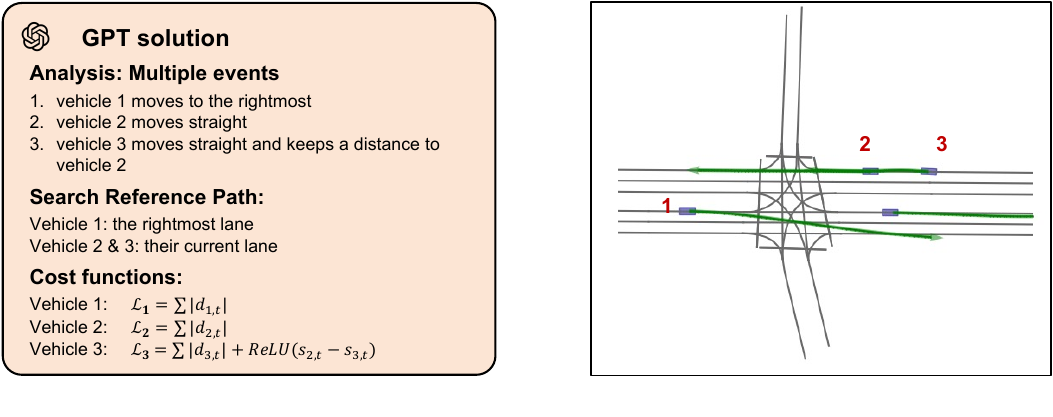}}
\\
\vspace{1.0em}
\subfloat[][\small{User query: vehicle 1 drives in reverse and out of the road, while vehicle 2 turns right.}] {\includegraphics[width=0.47\textwidth, trim={0 10.0 0 0cm}, clip]{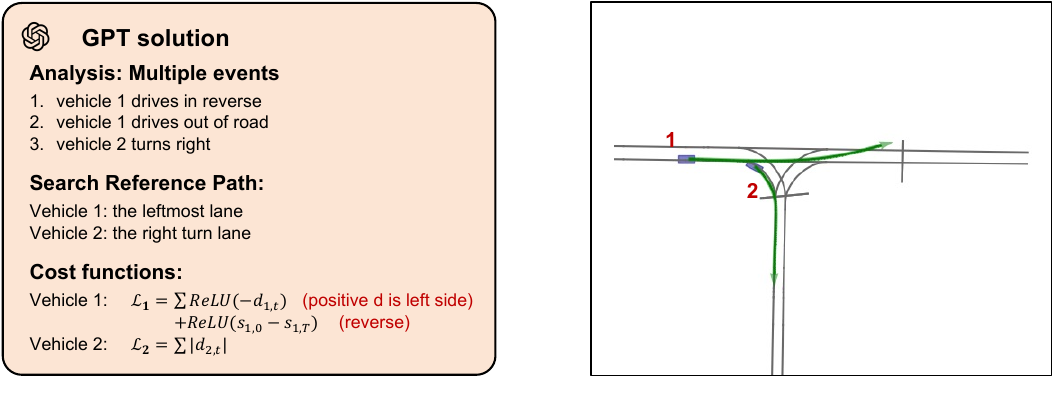}}
\hspace{0.03\linewidth}
\subfloat[][\small{User query: vehicle 1 drives left and right within its lane, and vehicle 2 keeps straight and follows vehicle 1.}] {\includegraphics[width=0.47\textwidth]{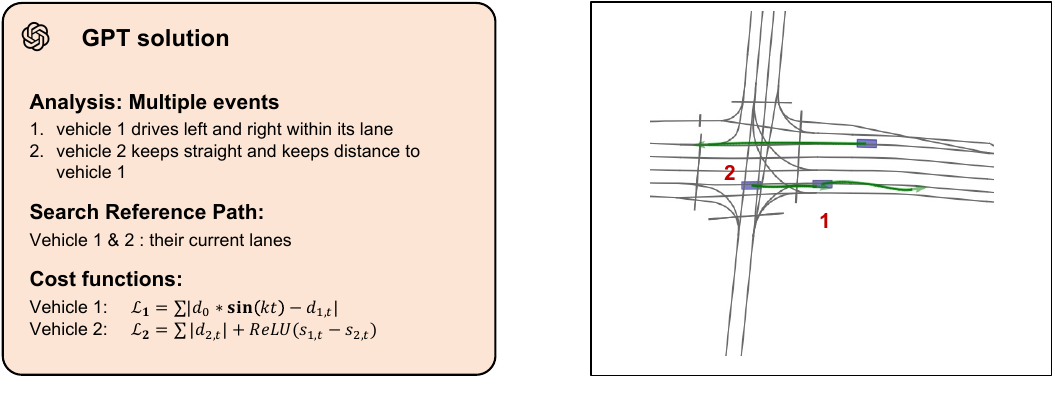}}
\caption{GPT solutions and visualizations for typical cases.}
\label{fig:vis}
\vspace{-4.0mm}
\end{figure*}

While our method demonstrates a high success rate, we observe that in some cases, the generated cost functions may contain errors or prove challenging for the diffusion model to follow due to different contexts in traffic scenarios. To address this, our refinement module plays a crucial role in improving the success rate, as evidenced by the results in Table~\ref{tab:succ_rate}. This module allows the LLM to adaptively adjust the cost functions, ensuring they better align with the intended guidance. Visualization examples in Figure~\ref{fig:refinement} further illustrate this process, showing how the refinement module effectively corrects and optimizes the cost functions, leading to a higher success rate and more accurate guidance.

\subsection{Visualizations}

We further combine some descriptions then query the LLM, and visualize the generation results, as shown in Fig.~\ref{fig:vis}. The LLM effectively decomposed the complex descriptions and generated the correct forms of cost functions. The visualizations further demonstrate that the model accurately follows the guidance of the generated functions.

\begin{figure}
    \centering
    \includegraphics[width = 0.45\textwidth]{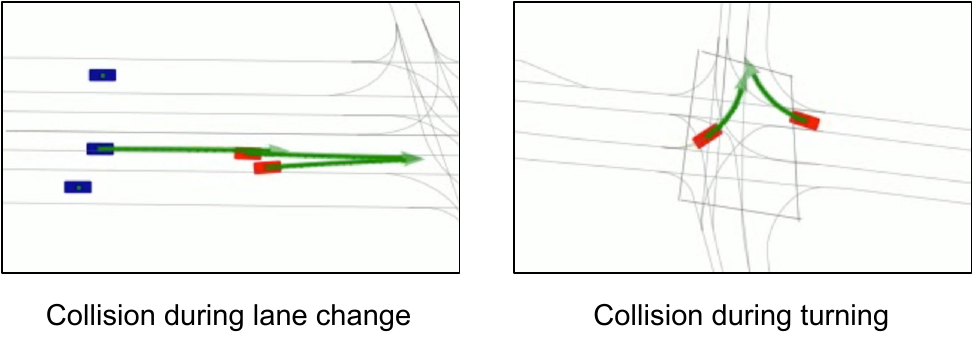}
    
    \caption{
    Visualizations of our generated accidents.
    }
    \label{fig:accident}
    \vspace{-6.0mm}
\end{figure}

Furthermore, our method demonstrates zero-shot capability in handling accident events, despite being trained exclusively on a nearly collision-free dataset, as illustrated in Fig.~\ref{fig:accident}. This ability is particularly valuable for traffic simulation, as it enables the model to generalize to rare but critical scenarios, such as accidents, without requiring explicit training data.

%% file: parts/06_conclusion.tex
\vspace{-1.0mm}

\section{Conclusion}\label{sec:conclusion}


In this paper, we proposed a novel LLM-enhanced controllable traffic simulation model. Our method is capable of taking users' descriptions as input, and generating a wide range of scenarios, including complex interactions and long-tail events. Several examples, including complex interactions and traffic accidents, have validated the effectiveness of our method. Finally, our numerous experiments and visualization demonstrate our framework achieves high-quality trajectory generation and precise instruction alignments across various benchmarks and settings.


\textbf{Limitations}. Although the refinement module effectively adjusts cost functions, it struggles to correct them when the LLM's initial understanding of the scenario is fundamentally incorrect. Future work could focus on enhancing the refinement module by incorporating more robust error detection mechanisms or leveraging external knowledge sources to improve its ability to correct fundamentally flawed initial assumptions.

